\documentclass[conference]{IEEEtran}

\usepackage[utf8]{inputenc} 
\usepackage[T1]{fontenc}    
\usepackage{url}            
\usepackage{booktabs}       
\usepackage{amsfonts}       
\usepackage{nicefrac}       
\usepackage{microtype}      
\usepackage{amsmath}
\usepackage{graphicx}
\usepackage{color}
\usepackage{pbox}
\usepackage[backend=bibtex, citestyle=numeric]{biblatex}
\title{LSTM with Working Memory}

\author{\IEEEauthorblockN{Andrew Pulver}
\IEEEauthorblockA{Department of Computer Science\\
University at Albany\\
Email: apulver@albany.edu}
\and
\IEEEauthorblockN{Siwei Lyu}
\IEEEauthorblockA{Department of Computer Science\\
University at Albany\\
Email: slyu@albany.edu}
}
\begin{document}
\maketitle

\begin{abstract}

Previous RNN architectures have largely been superseded by LSTM, or ``Long Short-Term Memory''. Since its introduction, there have been many variations on this simple design. However, it is still widely used and we are not aware of a gated-RNN architecture that outperforms LSTM in a broad sense while still being as simple and efficient. In this paper we propose a modified LSTM-like architecture. Our architecture is still simple  and achieves better performance on the tasks that we tested on. We also introduce a new RNN performance benchmark that uses the handwritten digits and stresses several important network capabilities.

\end{abstract}

\section{Introduction}

Sequential data can take many forms. Written text, video data, language, and many other forms of information are naturally sequential. Designing models for predicting sequential data, or otherwise extracting information from a sequence is an important problem in machine learning. Often, recurrent neural networks are used for this task. Unlike a non-recurrent network, a recurrent network's input at a given time-step consists of any new information along with the output of the network at the previous time-step. Since the network receives both new input as well as its previous output, it can be said to have a ``memory'', since its previous activations will affect the current output. Training a RNN model with gradient descent, or similar gradient-based optimization algorithms is subject to the usual problem of vanishing gradients \cite{bengio1994learning}. At each time step, the gradient diminishes and eventually disappears. Consequently, if the RNN needs information from more than a few time-steps ago to compute the correct output at the current time step, it will be incapable of making an accurate prediction. The model ``Long Short-Term Memory'', \cite{hochreiter1997long} greatly mitigated this problem. LSTM incorporates ``gates'', which are neurons that use a sigmoid activation, and are multiplied with the output of other neurons. Using gates, the LSTM can adaptively ignore certain inputs. LSTM also maintains a set of values that are protected by the gates and that do not get passed through an activation function.

In this work we develop a modification to LSTM that aims to make better use of the existing LSTM structure while using a small number of extra parameters. We claim that there are three issues with LSTM with forget gates. First, forget gates impose an exponential decay on the memory, which may not be appropriate in some cases. Second, the memory cells cannot communicate or exchange information without opening the input and output gates, which also control the flow of information outside the memory cells. Third, the hyperbolic tangent function is not ideal since LSTM memory values can grow large, but the the hyperbolic tangent has a very small gradient when its input value is large.

In our modified version, the forget gate is replaced with a functional layer that resides between the input and output gates. We call this modification LSTM with working memory, henceforth abbreviated LSTWM. LSTWM incorporates an extra layer that operates directly on the stored data. Rather than multiplying previous cell values by a forget gate, it uses a convex combination of the current memory cell value and the output of this extra layer whose input consists of previous memory cell values. In addition, we find that using a logarithm-based activation function improves performance with both the LSTM architecture and our modified variant. We evaluate this method on the Hutter challenge dataset \cite{hutter} as well as a task designed using the MNIST \cite{lecun2010mnist} handwritten digit dataset.

\section{Architecture and Training}

We begin with a review of LSTM. Since nearly all modern implementations use forget gates, we will call LSTM with forget gates standard LSTM or just LSTM. The term ``memory cell'' will be used to refer to the inner recurrence with a weight of one and ``LSTM cell'' to refer to an entire, individual LSTM unit. 

\subsection{Background on LSTM}

We introduce notation to be used throughout the rest of the paper: $x_t$ is the input vector at time $t$, $y_t$ is the network layer output at time $t$, $\sigma=\frac{1.0}{1.0+e^{-x}}$ and $f$ is an activation function. The standard LSTM formulation includes an input gate, output gate, and usually a forget gate (introduced in \cite{gers2000learning}). The output of an LSTM unit is computed as follows:

\begin{align}
a &= f(W\cdot [x_t;y_{t-1}]+b) \\
g^{(i)} &= \sigma(W_{g_i}\cdot [x_t;y_{t-1}] + b_{g_i}) \\
g^{(o)}&= \sigma(W_{g_o}\cdot [x_t;y_{t-1}] + b_{g_o}) \\
g^{(f)} &= \sigma(W_{g_s}\cdot [x_t;y_{t-1}] + b_{g_s}) \\
c_t &= g^{(i)}\odot a + g^{(f)}\odot c_{t-1} \\
y_t &= g^{(o)}\odot f(c_t)
\end{align}

\begin{figure}[t!]
  \centering
  \includegraphics[width=.4\textwidth]{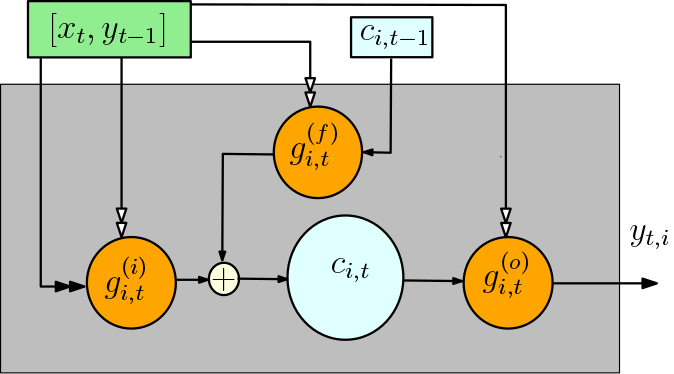}
  \caption{LSTM with forget gates. The orange circles are multiplicative gate units, the white circle is the addition operator. Double arrows represent the output of a neuron with multiple inputs, while single arrows represent the path of a single value.}
  \label{fig:lstmfig}
\end{figure}

It should be noted that other variants of LSTM exist, such as LSTM with peepholes \cite{gers2003learning} and more recently Associative LSTM \cite{danihelka2016associative}. We chose LSTM with forget gates as it is a simple yet commonly used LSTM configuration. 

The input and output gate values, $g^{(i)}$ and $g^{(o)}$, serve to regulate incoming and outgoing data, protecting the memory cell value. The forget gate can be used to clear the memory cell when its data is no longer needed. If the input/output gates are closed, and the forget gate is open, then there is no transformation on the memory cell values. This is an extreme  example, since the gate cells will likely take different values at any given timestep and can never be fully open or closed. However, it illustrates the main point of LSTM, which is that it can store data for an arbitrary amount of time, in principle.

LSTM with forget gates does not share the same mathematical properties of the original LSTM design. The memory-cell value decays exponentially due to the forget gate. This is important, since the network has a finite memory capacity and the most recent information is often more relevant than the older information. It may, however, be useful to decay that information in a more intelligent manner. Another point is that the information in the memory cells cannot be used without releasing it into the outer recurrence. This exposes them to the downstream network. A memory cell must also, to some degree, accept information from the upstream network to perform computation on the recurrent information.

Another related architecture in this domain is the Gated Recurrent Unit, or GRU \cite{cho2014learning}, which uses a convex combination instead of an input gate. It does not use an output gate. This makes the architecture easier to implement, since there is only one set of recurrent values per layer. This also does not share the theoretical properties of the original LSTM, since the memory cell values are replaced by their input rather than summed with their input.

Empirical studies \cite{jozefowicz2015empirical} \cite{chung2014empirical} comparing the performance of LSTM and GRU are somewhat inconclusive. Despite this, we hypothesize that a potential weak-point of the GRU is that its cells cannot accumulate large values as evidence of the strong presence of a certain feature. Suppose a GRU sees the same feature three times in a row, and its input gate is open. The cell will (roughly speaking) be replaced three times, but there is comparatively little evidence that the GRU saw this feature more than once. On the other hand, the LSTM memory cell has been increased three times, and it can arguably retain this large-magnitude value for a longer period of time. For this reason, we feel that the LSTM is a stronger base upon which to build.

\begin{figure}[t!]
  \centering
  \includegraphics[width=.4\textwidth]{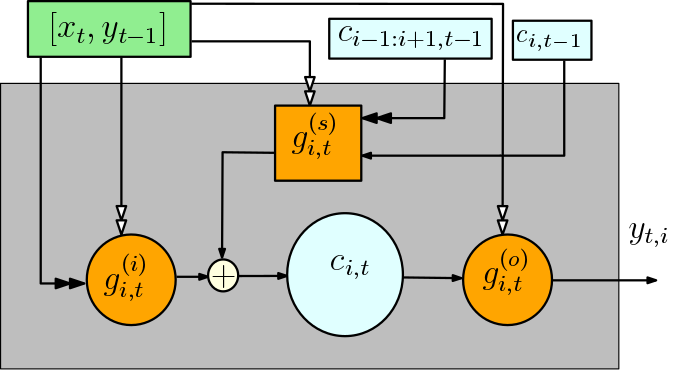}
  \caption{LSTWM. The orange square represents a convex combination of the two inputs on the right.}
  \label{fig:lstwmfig}
\end{figure}

\subsection{LSTWM}

For a regular LSTM to operate on the values stored in its memory cell and store them again, it must open both the output gate and the input gate. This releases the value to the downstream network, and new input from the upstream network must also, to some degree, be allowed to enter the cell. We propose that a network may make better use of its input and output gates if they weren't serving the two, not necessarily related, purposes of regulating recurrent data as well as regulating inputs and outputs to and from other parts of the network. Additionally, the activation function is applied to the memory cell values before they are multiplied by the output gate. Typically, the hyperbolic tangent function is used, and if the memory cell value already has a high magnitude, the errors entering the memory cell will be very small. We argue that a non-saturating activation function will confer a significant performance increase.  The output of a LSTWM cell is computed in the following manner:

\begin{align}
a &= f(W\cdot [x_t;y_{t-1}]+b) \\
g^{(i)} &= \sigma(W_{g_i}\cdot [x_t;y_{t-1}] + b_{g_i}) \\
g^{(o)}&= \sigma(W_{g_o}\cdot [x_t;y_{t-1}] + b_{g_o}) \\
g^{(s)} &= \sigma(W_{g_s}\cdot [x_t;y_{t-1}] + b_{g_s}) \\
c^{(r_l)} &=  \rho(c_{t-1},-1) \\
c^{(r_r)} &=  \rho(c_{t-1},1) \\
i_t &= f (w_{v_1}\odot c_{t-1} + w_{v_2} \odot c^{(r_l)} +  w_{v_3} \odot c^{(r_r)} + b_{v_1}) \\
r_t &= g^{(s)} \odot c_{t-1} + (1.0-g^{(s)}) \odot i_t \\
c_t &=  g^{(i)}\odot a + r_t \\
y_t &=  g^{(o)}\odot f(c_t)
\end{align}

Where $f$ is an activation function, either

\[ f(x) = 
\begin{cases}
-ln(-x+1) &  \text{if } x \text{< 0} \\
ln(x+1) &  \text{otherwise}
\end{cases}
\]

or
\[ f(x) = 
tanh(x)\\
\]

\begin{figure}[t!]
  \centering
  \includegraphics[width=.5\textwidth]{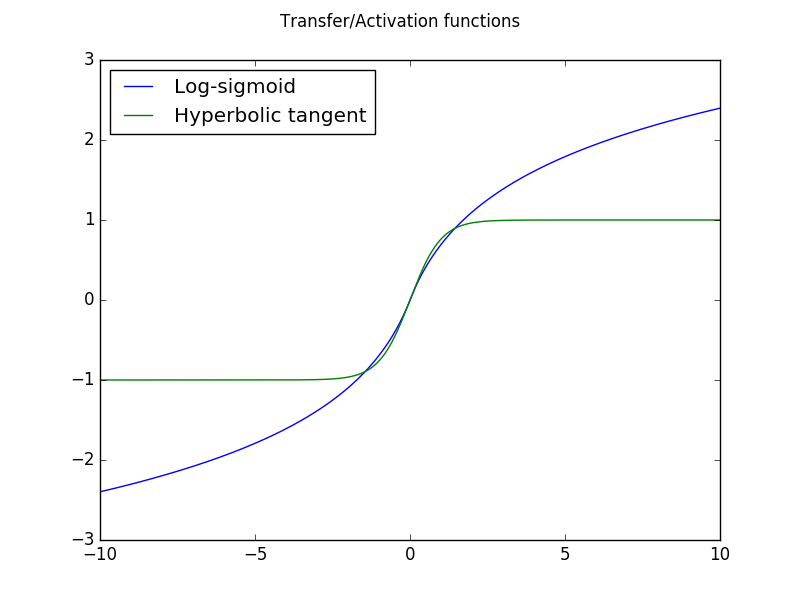}
  \caption{Comparison of tanh and log-based activation function}
  \label{fig:activation}
\end{figure}

This logarithm-based activation function does not saturate and can better handle larger inputs than tanh. Fig. ~\ref{fig:activation} illustrates both functions. The recurrent cell values may grow quite large, causing the hyperbolic tangent function to quickly saturate and gradients to disappear. To obtain good performance from our design, we wanted to develop a non-saturating function that can still squash its input. Earlier works have used logarithm-based activations, and the function we used appears in \cite{isa2010suitable} and originally in \cite{bilski2000backpropagation}.

While the rectified linear unit \cite{glorot2011deep} (ReLU) is a common choice for non-recurrent architectures, they are usually not suitable for LSTM cells since they only have positive outputs (although they have been used with some success in LSTM and RNNs, see \cite{le2015simple} \cite{krueger2015regularizing}), and exploding feedback loops are more likely when the activation function does not apply some ``squashing effect''. However, tanh is less suited for large inputs due to the small values of its derivative outside the small interval around zero. 

\begin{figure}[t!]
  \centering
  \includegraphics[width=.5\textwidth]{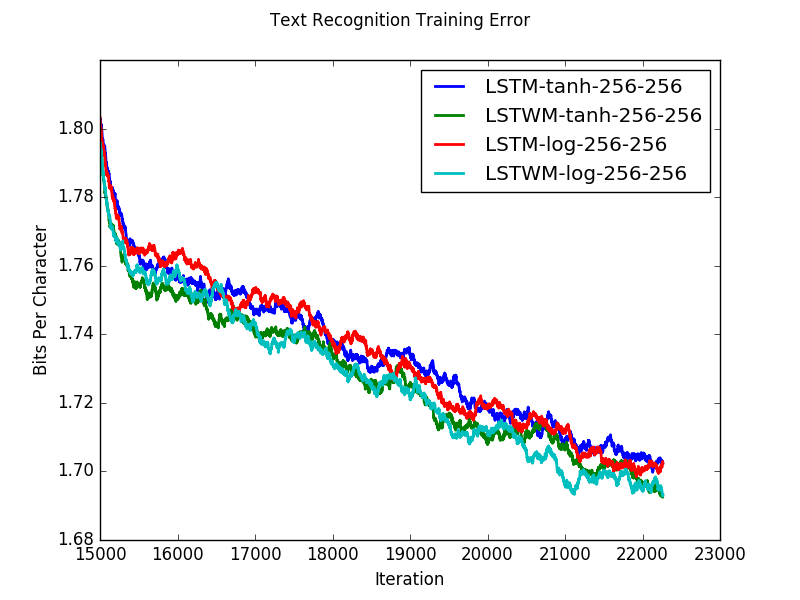}
  \caption{Wikipedia Text Prediction Task. Two layer networks.}
  \label{fig:text_training_2l}
\end{figure}

The forget gate output has been renamed $g^{(s)}$, and now controls the convex combination of the previous memory cell value $c_{t-1}$ with the output of the inner layer $i_t$.  The weight vectors $w_{v_2}$ and $w_{v_3}$ only connect a neuron to its left and right neighbors. In software, we did not implement this as a matrix multiplication, but as an element-wise roll of the input which gets multiplied by weight vectors and then has bias vectors added. The function $\rho(x,y)$ is an element-wise roll where $x$ is the input vector and $y$ is the number of times to roll left or right. For example $\rho([1,2,3],1) = [3,1,2]$ and  $\rho([1,2,3],-1) = [2,3,1]$. So the inter-neuron connections within the memory cells are sparse and local. The reason for choosing a sparse layer is that dense layers grow quadratically with respect to layer width. We wanted to compare equal-width networks, since a LSTM's memory capacity is determined by the number of individual memory cells. By setting near-zero weights and biases for the inner layer, the network can also achieve a traditional forget gate. Since this layer does not use many extra parameters, we can compare equal width networks. $w_{v_1}$, $w_{v_2}$, $w _{v_3}$ and $b_{v_1}$ are initialized with zeros, so the network starts out as a standard LSTM. In summary, an LSTWM network can modify its memory in a more complex manner without necessarily accepting new values or exposing its current values. Since the inner layer only uses the previous memory cell values, it can be computed in parallel with any downstream network and does not present a computation bottleneck if implemented in a parallel manner.

This architecture was adapted from a design in a previous version \cite{pulver2016lstm} of this work, which used normal forget gates and included an extra layer and extra gate after the memory cell update. We found that the previous four-gate architecture did not perform as well on tasks that required a precise memory. Likely, having three different memory operations at each timestep resulted in excessive changes in to the memory. Setting appropriate initial bias values helped the situation in some cases, however, we found better designs that did not require as much hand-tuning. Our first attempt at resolving the issue was removing the forget gate. Removing the forget gate from our earlier design did not yield good result by itself.

Figures  ~\ref{fig:lstmfig} and ~\ref{fig:lstwmfig} illustrate LSTM and LSTWM, respectively. The $i$ subscript indicates that this is the $i^{th}$ unit in a layer. The white double arrows indicate an input that determines a gate value (as opposed to a value that gets multiplied by a gate).

\begin{figure}[t!]
  \centering
  \includegraphics[width=.5\textwidth]{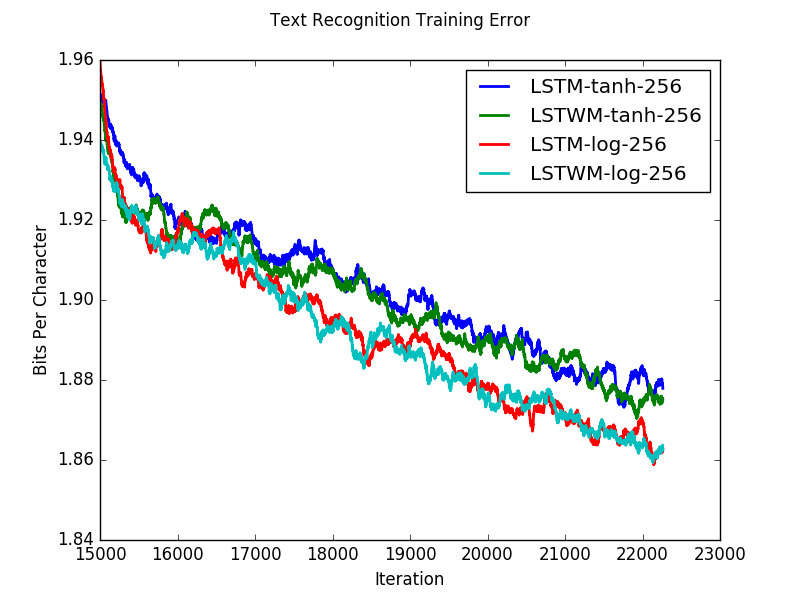}
  \caption{Wikipedia Text Prediction Task. Single layer networks.}
  \label{fig:text_training_1l}  
\end{figure}

\subsubsection{Training, Regularization and Other Details}

Given that the memory cells are not necessarily decaying in an exponential manner at each timestep, it is important to control their magnitude. Rather than just using a regularization term on the network weights themselves, we also regularize the overall magnitude of the memory-cells at each timestep. 

When training an ANN of any type, a regularization term is usually included in the cost function. Often, the L2-norms of the weight matrices are added together and multiplied by a very small constant. This keeps the weight magnitudes in check. For our architecture to train quickly, it is important to use an additional regularization term. Given a batch of training data, we take the squared mean of the absolute value plus the mean absolute value of the memory cell magnitudes at each timestep for every element in the batch. In other words, $\eta\cdot(mean(|cells|)^2 + mean(|cells|))$ for some small constant $\eta$. We found that using $\eta\approx 10^{-2}$ or $\eta\approx 10^{-3}$ worked well. Similar regularization terms are discussed in \cite{krueger2015regularizing}, although they differ in that they are applied to the change in memory cell values from one timestep to the next, rather than their magnitude. Using direct connections between the inner cells can amplify gradients and produce quickly exploding values. By adding this regularization term, we penalize weight configurations that encourage uncontrollable memory cell values. Since the cells values get squashed by the activation function before being multiplied by the output gate, regularizing these values shapes the optimization landscape in a way that encourages faster learning. This is also true to some extent for regular LSTM, which benefits from this type of regularization as well. We also applied this regularization function to the weights themselves. LSTWM can learn effectively without extra regularization, but it will learn slowly in some cases.

Note that we use the square-of-abs-mean not the mean-squared. Using the square-of-abs-mean allows some values to grow large, and only encourages most of the values to be small. This is important, especially for the memory cells, because the ability of a neural network to generalize depends on its ability to abstract away details and ignore small changes in input. Indeed, this is the entire purpose of using sigmoid-shaped squashing functions. If the goal were to keep every single memory cell within tanh's ``gradient-zone'', one could use the hyperbolic cosine as a regularization function since it grows very large outside this range. 

We used Python and Theano \cite{2016arXiv160502688short} to implement our network and experiments.

\section{Experiments}

We have two experimental tasks to test our network on: text prediction and a combination digit-recognition and addition task. The networks were trained using ADAM \cite{kingma2014adam}, an optimization algorithm based on gradient descent, with the following settings $\alpha=.001$ $\beta_1=.9$ $\beta_2=.999$.

\section{Text Prediction}

\begin{figure}[t]
  \centering
  \begin{tabular}{|c|c|c|c|c|}
    \hline
    Architecture & BPC on test set after training \\
    \hline
    LSTM-256-tanh & 1.893 \\
    \hline
    LSTWM-256-tanh & 1.892\\
    \hline
    LSTM-256-log & 1.880 \\
    \hline
    LSTWM-256-log & 1.880 \\
    \hline
    LSTM-256-256-tanh & 1.742 \\
    \hline
    LSTWM-256-256-tanh & 1.733 \\
    \hline
    LSTM-256-256-log & 1.730 \\
    \hline
    LSTWM-256-256-log & 1.725 \\
    \hline    
  \end{tabular} 
  \caption{Performance on text prediction task.}
  \label{fig:text_results}
\end{figure}

Like many other works e.g. \cite{danihelka2016associative}, we use the hutter challenge \cite{hutter} dataset as a performance test. This is a dataset of text and XML from Wikipedia. The objective is to predict the next character in the sequence. (Note that we only use the dataset itself, and this benchmark is unrelated to the Hutter compression challenge) The first 95\% of the data was used as a training set and the last 5\% for testing. Error is measured in bits-per-character, BPC, which is identical to cross entropy error, except that the base-2 logarithm is used instead of the natural log. We used a batch size of 32 and no gradient clipping. To reduce the amount of training time needed, we used length-200 sequences for the first epoch, and then used length-2000 sequences for the remaining five epochs.

Figures ~\ref{fig:text_training_2l} and ~\ref{fig:text_training_1l} show a running average of the training error and ~\ref{fig:text_results} shows the BPC on the test set after training. The results are close for this particular task, with LSTWM taking a slight advantage. Notice that for this test, the logarithm based activation does carry some benefit, and the best performing network was indeed LSTWM with the logarithmic activation.

\subsection{Training Information}

Given the popularity of this task as a benchmark, a quick training phase is desirable. Input vectors are traditionally given in a one-hot format. However, the network layers can be quite wide, and each cell has many connections. We found that using a slightly larger nonzero value in the vectors resulted in quicker training. Instead of using a value of 1.0 for the single non-zero element in the input vectors, we used $log(n)+1.0$ where $n$ is the number of input symbols. In this case, $n=205$, since there are 205 distinct characters in this dataset. On tasks where there are many symbols, the mean magnitude of the elements of the input vector is not as small, which accelerated training in our experiments. 

Another method we used to speed up training was using a pre-train epoch with shorter sequences. The shorter sequences mean that there are more iterations in this epoch and the iterations are performed more quickly. This rapid pre-training phase moves the parameters to a better starting point more quickly. This method could also be used in any task where the input is an arbitrary-length sequence. Longer sequences result in better quality updates, but they take proportionally longer to compute. We also added a very small amount of Gaussian noise to the inputs during the pre-train epoch, but not during the main training phase. Fast and noisy updates work well early in the training process and reduce the amount of time necessary to train networks.

\section{Digit recognition and addition combined task}
\begin{figure}[t]
  \centering
  \includegraphics[width=.3\textwidth]{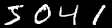}
  \\
  \vspace{.1cm}
  \includegraphics[width=.3\textwidth]{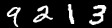}
  \\
  \vspace{.1cm}
  \includegraphics[width=.3\textwidth]{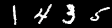}
  \caption{Example input sequences for combination digit task}
\end{figure}

\begin{figure}[t]
  \centering
  \includegraphics[width=.4\textwidth]{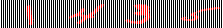}
  \caption{Inputs colored and separated into columns to show individual input columns}
\end{figure}

\begin{figure}[t]
  \centering
  \includegraphics[width=.5\textwidth]{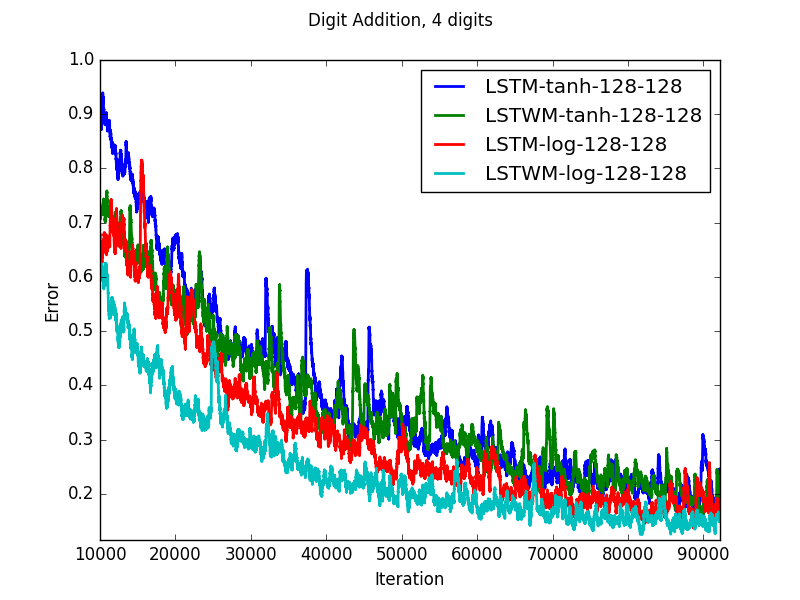}
  \caption{Digit Combo task training error.}
  \label{fig:combo_training}    
\end{figure}

\begin{figure}[t!]
  \centering
  \begin{tabular}{|c|c|}
    \hline
    Architecture & \pbox{20cm}{ \# of correct sum outputs \\ \centering{(out of 10000)} } \\
    \hline
    LSTM-128-128-tanh & 8864 \\
    \hline
    LSTWM-128-128-tanh & 8820 \\
    \hline
    LSTM-128-128-log & 8972 \\
    \hline
    LSTWM-128-128-log & 9015 \\
    \hline
  \end{tabular} 
  \caption{Performance on digit combo task.}
  \label{fig:combo_results}  
\end{figure}

This task is a combination of digit recognition and addition. Other works (e.g. \cite{ha2016hypernetworks}, \cite{kalchbrenner2015grid}) use the MNIST dataset as a performance test for artificial neural networks. Addition is also a common task for evaluating the viability of new RNN designs, and usually the numbers to be added are explicitly encoded by a single input vector at each timestep. The difference with this task is that we horizontally concatenate a number of images from MNIST and train on the sum of those digits. This large concatenated image is presented to the network in column vectors. After the network has seen the entire image, it should output the sum of the digits in the input. No error is propagated until the end of the image. This makes the task a useful benchmark for memory, generalization ability, and vision. The training targets are placeholders until the end, and then two (more if needed) size-10 vectors corresponding to the two digits of the sum. A similar but non-sequential task appears in \cite{ba2014multiple}.

For this task, the error coming from the end of the sequence must be back-propagated all the way to the first digit image. If this does not occur, the model will not correctly learn. Unlike text prediction, this requires a robust memory and the ability to back-propagate the error an arbitrary number of time-steps. We suspect that text prediction is a task where memory values are often replaced.  For text-prediction, at the end of training, the forget gate biases are often < 0, which indicates low forget-gate activity, i.e. it tends to forget things quickly. This task was developed as a way to test memory functionality, since text-recognition does not stress this capability. 

Another interesting property of this task is that it demonstrates the ability of an architecture to form a generalized algorithm. Increasing the length by concatenating more digit images will remove non-algorithmic minimums from the search space. In other words, the network will need to learn an addition algorithm to correctly process a longer sequence. Simply ``memorizing'' the correct outputs will not work when the network does not have enough parameters, to work on longer sequences, it must learn an addition algorithm. This can also be applied without the strict memory requirements of back-propagating to the first digit. In this case, one would simply train to output the running sum at frequent intervals rather than just at the end of the sequence. In the boundary case where the number of digit images is one, it becomes a simple digit-recognition task. 

We trained in a similar manner to the text-prediction experiment. Each training input consists of randomly selected digits along with their sum as a target. The same ADAM parameters were used as in the text-prediction task. We trained on sequences with four concatenated digit images. Therefore, each input sequence was 112 input vectors with three more placeholders at the end. As a preprocessing step, the digit images were added with Gaussian (scaled down $\times 10^{-5}$) noise, blurred (3x3 kernel, Gaussian), mean-subtracted and squashed with the log-activation function in that order. Better performance can be obtained by omitting the noise and blur, however it was difficult to obtain consistent results without some smoothing of the inputs.

The LSTWM architecture learned more quickly, and achieved the best performance, although only using the logarithm-based activation. Figures ~\ref{fig:combo_training} and ~\ref{fig:combo_results} show the training error curve and peak test set results respectively.

\subsection{Plain Digit Recognition}

We also tested the case for the task above where there is only one digit image, for a variety of different network sizes. Since the layers are narrow, we decided to slightly increase the width of LSTM so that the LSTM networks have more parameters. Since this is a much simpler task, it did not require as much training.

\begin{figure}[t!]
  \centering
  \includegraphics[width=.5\textwidth]{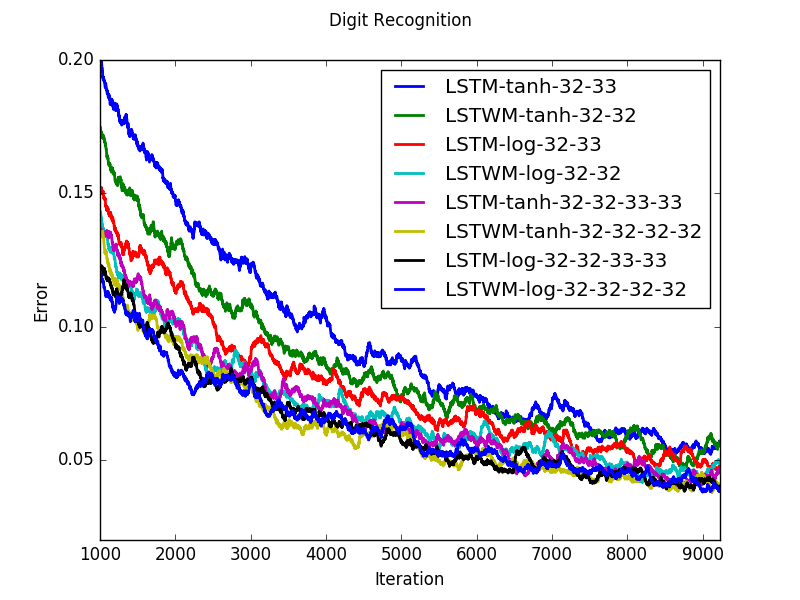}
  \caption{Plain Digit Recognition}
  \label{fig:digit_training}    
\end{figure}

\begin{figure}[t!]
  \centering
  \begin{tabular}{|c|c|}
    \hline
    Architecture & \pbox{20cm}{ Best test-set performance \\ \centering{(out of 10000)} } \\
    \hline
    LSTM-32-33-tanh & 9600 \\
    \hline
    LSTWM-32-32-tanh & 9620 \\
    \hline
    LSTM-32-33-log & 9657 \\
    \hline
    LSTWM-32-32-log & 9655 \\
    \hline
    LSTM-32-32-33-33-tanh & 9676 \\
    \hline
    LSTWM-32-32-32-32-tanh & 9690 \\
    \hline
    LSTM-32-32-33-33-log & 9690 \\
    \hline
    LSTWM-32-32-32-32-log & 9712 \\
    \hline 
  \end{tabular} 
  \caption{Performance on digit recognition. (out of 10,000 test sequences)}
  \label{fig:digit_results}      
\end{figure}

\subsection{Results}

For text-prediction, the performance is roughly equal, although LSTWM with the logarithm activation slightly outperforms the other networks. Notably, the logarithmic activation function increases performance consistently on this task. For the digit-combo task, we observed better performance with LSTWM, and likewise on the simple digit recognition task. The number of extra parameters used is linear, not quadratic, with respect to layer width, so we consider it a good improvement given the small size increase. We claim that using the inner layer, along with using a logarithm-based activation will offer a modest but significant performance benefit. Another notable feature of LSTWM is that the training error decreases more quickly compared to standard LSTM.

\section{Conclusion}

We discussed several interesting properties of LSTM and introduced a modified LSTM architecture that outperforms LSTM in several cases using few additional parameters. We observe that a logarithm-based activation function works well in LSTM and our modified variant. Finally, we presented a benchmark for recurrent networks based on digit-recognition.

\printbibliography

\end{document}